\documentclass[a4paper]{article}

\usepackage{INTERSPEECH2020}
\usepackage{tipa}
\usepackage{xcolor}
\usepackage{multirow}
\usepackage{bbm}
\title{Perceptimatic:\\A human speech perception benchmark for unsupervised subword modelling}
\name{Juliette Millet$^{1,2,3}$, Ewan Dunbar$^{1,2}$}
\address{$^1$ CoML, ENS/CNRS/EHESS/INRIA/PSL Research University, Paris, France\\$^2$ Universit{\'{e}} de Paris, LLF, CNRS, Paris, France\\$^3$CRI, Département Frontières du Vivant et de l'Apprendre, IIFR, Universit{\'{e}} de Paris}
\email{juliette.millet@cri-paris.org}

\begin{document}

\maketitle
\begin{abstract}
 In this paper, we present a data set and methods to compare speech processing models and human behaviour on a phone discrimination task. We provide Perceptimatic, an open data set which consists of French and English speech stimuli, as well as the results of 91 English- and 93 French-speaking listeners. The stimuli test a wide range of French and English contrasts, and are extracted directly from corpora of natural running read speech, used for the 2017 Zero Resource Speech Challenge. We provide a method to compare humans' perceptual space with models' representational space, and we apply it to models previously submitted to the Challenge. We show that, unlike unsupervised models and supervised multilingual models, a standard supervised monolingual HMM--GMM phone recognition system, while good at discriminating phones, yields a representational space very different from that of human native listeners.
\end{abstract}

\section{Introduction}

Not all errors are equal. If your task is to identify animals, mistaking a chimpanzee for a marmoset is less important than mistaking a chimpanzee for a hippo. This is also true when evaluating models that deal with human speech. Speech sounds differ from each other in different ways, and mistaking one sound for another in a given language is less important if the two sounds are also highly confusable for human native speakers.

Several works have  made detailed comparisons between human speech perception experiments and automatic speech recognition (ASR) and related systems. In \cite{stolcke2017comparing},   neural ASR  word transcription errors  were compared qualitatively with errors made by human annotators, 
while \cite{meyer2007phoneme} compared HMM phoneme confusions with the results of a human  phoneme labelling experiment. 
The finer-grained study in \cite{pinter2016gmm} compared Japanese speakers' perceptual boundaries for the (allophonic) [s]/[\textctc] contrast with the behaviour of  GMM phoneme classifiers, while \cite{scharenborg2018visualizing} compared phonetic adaptation by Dutch listeners to artificially ``accented'' productions of the [r]/[l] to  acoustic models adapted to the experimental stimuli. Several studies also investigate  whether  ASR and unsupervised  speech representation learning accords with established phenomena in human speech perception at a qualitative level \cite{feldman2007rational,schatz2019early,schatz2017asr,thomashmm}. 

All of these comparisons focus either on one language, one model, or a handful of sounds. Most deal with models that predict phoneme labels, using methods that do not work for unsupervised models. In this paper, we present a large-scale phone discrimination task that can be performed by a wide range of models, and make a quantitative comparison with human listeners performing the same task.

Our contribution is twofold. First, we provide \textbf{Perceptimatic}, a data set containing stimuli and human results for a speech perception task in English and French (an extension of the English-only data described in \cite{cogsci2020}). Second, we use this data to evaluate the models submitted to the 2017 Zero Resource Speech Challenge (ZeroSpeech: see below).

We base our data set on test data used in the evaluation of the ZeroSpeech Challenge \cite{versteegh2015zero,dunbar2017zero,dunbar2019zero}. This machine learning challenge series aims to find methods for autonomously acquiring language, focusing on unsupervised learning for speech-related tasks. We focus on the French and English unsupervised subword modeling task in the 2017 edition. This task proposes to build systems that autonomously learn phoneme-like representations of speech from raw, continuous speech recordings. It is evaluated by a global phone discriminability measure applied to the learned representations. We provide a clean and manually verified subset of the 2017 test stimuli, along with experimental results from French- and English- speaking listeners on the phone discrimination task. To our knowledge, \textbf{Perceptimatic} is the first freely available data set of stimuli taken from natural, running French and English speech accompanied with human responses in a phone discrimination task, and as such, is the first  human data comparable to the automatic phone discriminability evaluation used for unsupervised subword modelling.\footnote{All  code used for the experiments, as well as stimuli and human results are available online at: https://github.com/JAMJU/interspeech-2020-perceptimatic, an extension of https://github.com/JAMJU/Cogsci2020-Perceptimatic-English} 

We use two methods to apply these results to the models submitted to ZeroSpeech 2017: we comparing models on how well they predict human behaviour in a speech discrimination task using the method in \cite{millet2019comparing}, and, in addition, we re-weight the results of the original ZeroSpeech 2017 phone discriminability scores to take account of native listeners' relative discriminability of different sounds.  We show that, with this reweighting, one of the submitted systems performs better than the challenge's supervised ASR reference system. We also show that this supervised ASR system, while performing well on the phone discriminability evaluation, has a representational space that is very different from that of human native listeners, and very different from the systems submitted to the ZeroSpeech challenge.



\section{Human and machine discrimination}
\label{sec:track1}
\subsection{Machine ABX discrimination}

A machine ABX phone discrimination test is a binary decision task in which two speech stimuli, A and B, which only differ in their centre phone (for example, [\textipa{seIk}]--[\textipa{soUk}] or [\textipa{zfA}]--[\textipa{zpA}]), are presented. A third stimulus, X, must be identified as being more similar to either A or B. This stimulus also shares the same first and last phones, and has the same centre phone as either A or B. $R_A$, $R_B$ and $R_X$, the representations of respectively A, B, and X, are extracted from the models to be evaluated, and distances $d(R_A,R_X)$ and $d(R_B,R_X)$ are computed. Machine ABX phone discrimination tasks are often used for evaluating unsupervised speech models. They are generic, and are also appropriate for evaluating supervised models.

Since 
stimuli are not all of the same duration, distance functions based on dynamic time warping are typically used (notably in the ZeroSpeech Challenges). Dynamic time warping takes two  sequences $C$ and $D$ as input (in our case either $R_A$ and $R_X$ or $R_B$ and $R_X$), as well as a function $\gamma$ for comparing pairs of sequence elements. It aligns  $C$ and $D$ by matching the elements of one to the other so as to minimize the sum of $\gamma(c, d)$ for all matched elements $(c,d)$. Each element of $C$ must be matched with at least one element of $D$, and alignments must respect temporal order. We calculate distances between stimuli $C = {c_1, c_2,...c_p}$ and $D = {d_1, d_2, ... d_q}$ as:
\begin{equation}
    d(C,D) = \frac{\sum_{c_i, d_j \text{ are matched }} \gamma(c_i, d_j)}{max(p, q)}
\end{equation} 
As in ZeroSpeech 2017, we take $\gamma$ as either the arc cosine of the normalized dot product or the symmetrised KL-divergence.

%
Once $d(R_A,R_X)$ and $d(R_B,R_X)$ are obtained using one of these methods, we compute $\delta = d(R_B,R_X) - d(R_A,R_X)$ if A and X belong to the same category, $\delta = d(R_A,R_X) - d(R_B,R_X)$ if it matches $B$. If $\delta>0$, then the model is considered to be correct, otherwise, it is considered to be wrong. We perform this operation on many ABX triplets for a given pair of phones. The percent accuracy gives a measure of the model's discriminability of the two phone categories.

The ZeroSpeech 2017 evaluation is an ABX phone discrimination task in which the A, B, and X stimuli are extracts taken from running speech. Models generate representations of audio files containing stimuli plus a surrounding context, from which their representations of the stimuli are extracted. In the current paper, we focus on the condition in which models are given one-second audio files, and in which X is uttered by a different speaker than A and B (across speaker). The ZeroSpeech 2017 evaluation measure was the average accuracy over all phone pairs.\footnote{In \cite{dunbar2017zero}, this average was computed in several steps by averaging accuracies the across contexts (flanking phones), then across speakers, and then across all centre phones. In our case, the test set on which we evaluate the models is almost exactly balanced for context, speakers, and centre phones. Thus, when we calculate accuracies, we simply take the global accuracy over all triplet items.}
Instead of relying only on this accuracy, we propose to evaluate models with respect to a human reference data set.

\subsection{Human ABX discrimination}

An ABX phone discrimination task for human participants is as follows: participants hear three stimuli in sequence (a triplet) and are asked to identify which one of the first two stimuli (A or B) is more similar to the third (X). As in the machine ABX task, the stimuli are such that one of the two responses is always considered correct (here, the centre phone of X matches either that of A or of B). We use the same stimuli as for the machine ABX task, and obtain multiple human responses for each triplet, which we code as correct or incorrect. 
We take two approaches to dealing with this data. First, we combine data across participants to compute an item-level accuracy for each triplet item. These item-level accuracies can the be averaged into accuracies for individual phone contrasts, comparable with those calculated in the machine ABX task. Alternatively, we can drop the notion of category discriminability, and  compare models' gradient  $\delta$ values 
with the probability that  listeners give the correct answer, at the level of  individual triplet items.

\subsection{Comparing model and human performance}
\label{sec:method_comp}

We relate human and model results in two ways. First, in order to have a phone discrimination score that takes into account  differences in how ``hard'' each triplet item is, we weight models' decisions by human accuracies as follows:

\begin{align}
\label{eq:rew}
    RewAcc_{ABX} = \frac{\sum_{t\in test} \mathbbm{1}_{\delta_M(t) > 0} \times Hum(t)}{\sum_{t\in test} Hum(t)}
\end{align}
where $\delta_M(t)$ is the $\delta$ value for a triplet item $t$ given by a model $M$, and with $Hum(t)$ is the percent accuracy for human listeners for the triplet item $t$ (for French items, the accuracy is computed over French listeners, and for English items, the accuracy is computed over English listeners). 
This score gives more importance to triplets that are ``easy'' for human native listeners, and thus more important to discriminate correctly. It reduces the impact of triplets that are ``hard'' for humans, either because A, B and X are perceived as very similar, or because X is perceived to be more similar to the ``wrong'' answer.

Second, we compare models' representational spaces with humans' perceptual space in a detailed way. To do this, we evaluate models' ability to predict individual human responses. We evaluate use the $\delta$ values as predictors in a binary regression, predicting whether a human listener will have the correct response or not on a given trial. Each model we evaluate is trained separately on French and on English, yielding two $\delta$ values.\footnote{For models  trained on other languages, we use the same $\delta$ values for French and English.}  For each model, we fit an overparameterized probit regression with two zero-one language indicators as bias predictors, one which is 1 for French observations, and another which is 1 for English observations. We then construct French- and English-only $\delta$ predictors by multiplying the two $\delta$ values by the two indicator variables.\footnote{We also add as a predictor a binary variable indicating whether the correct answer was presented first or second, and the trial's position in the experiment, as well as predictors for participant.}
We calculate the log-likelihood. Models with better (higher) log-likelihoods have representational spaces more similar to  humans', in the sense that relative distances in the model predict human discriminability.


\section{Perceptimatic data set construction}
 
 \label{sec:benchmark-stimuli}
The stimuli are taken from the French and English ZeroSpeech 2017 test stimuli. These stimuli  were originally generated using forced alignment on the LibriVox audio book collection, and identifying all sequences of three phones. There are several problems with these materials. First, the number of triplets is too large to feasibly test on a large number of subjects. Second, there are labelling errors, as well as phones which may not be part of the language variety of certain listeners. Third, the phonetic similarity of the centre phone between X and the correct answer is often doubtful because of contextual variability. Fourth, the flanking phones are sometimes very different for A, B and X, due to contextual variability. Finally, the phone boundaries are not precise enough.
We eliminated incorrectly labelled phones, and then, as native English and French listeners,  selected and realigned a subset of triplets by hand to minimize these issues.
In total, the cleaned subset consists of 5202 triplets (2214 from English), making 461 distinct centre phone contrasts (212 English, 249 French), in a total of 201 distinct contexts (118 English, 83 French), with most phone comparisons appearing in three contexts each (a total of 47 English contrasts appear in either one, two, or four contexts). The speakers used (15 English, 18 French) have, in our assessment, pronunciations close to standard American English/Metropolitan French.\footnote{The full set of English centre phones included in at least one item is $[$\textipa{\ae~A~b~d~D~eI~E~f~g~h~i I k l m n N oU p \*r s S t tS u U v \textturnv  w z}$]$. The full set of French phones is [\textipa{a \~A b d e E \~E f g i j k l m n \textltailn~o \o~O \~O p K s S t u v w y z Z}]. For the full list of pairs and contexts, see the online repository.}

The data set includes 91 participants located in the United States reporting English as the sole language to which they were primarily exposed up to the age of eight, and 93 located in France, attesting the same for French. They were recruited online on Amazon Mechanical Turk (all English speakers, and 57 French speakers) and in person, and all performed the test online with \textsc{LMEDS} software \cite{mahrt2013lmeds}.\footnote{\cite{kleinschmidt2015robust} compared data from an in-lab speech perception experiment with a Mechanical Turk replication and found a close correspondence. 48 additional participants (15 US, 33 France) were tested but did not meet the language background requirements, and 115 participants (65 US, 50 France) were rejected for failing at least three out of twelve catch trials or not finishing the task. The catch trials consisted of additional, highly distinct ABX stimuli, including several which required participants to distinguish \emph{cat} from \emph{dog} for English speakers or \emph{caillou} from \emph{hibou} for French speakers.} They were paid for participation. All participants were tested both on English and French stimuli. Here we only report results on English contrasts for English listeners and on French contrasts for French listeners.

For testing, triplets were counterbalanced into lists of 190 per participant.\footnote{No participant is tested twice on the same phone pair, and the combination of speakers is not predictive of the right answer.} Each triplet was tested three times, so that most contrasts are tested at least 36 times. Participants respond as to which of the two reference stimuli the probe corresponded to on a six-point scale, ranging from \emph{first for sure} to \emph{second for sure}, with two intermediate degrees of certainty for each. The data set includes both these responses and a binarized version. 
Except in Figure \ref{fig:results_plot}, we report only the binarized responses. 
%

\section{Experiments}
\label{sec:experiment}


We evaluate the models submitted to the 2017 ZeroSpeech challenge. We compare them with the topline representation used in the challenge, posteriorgrams from a supervised HMM-GMM phone recognition system with a bigram phone-level language model, trained with a Kaldi recipe \cite{povey2011kaldi}. We also compare them with the baseline representation, mel filterbank cepstral coefficients (MFCCs) (thirteen first coefficients with $\Delta$ and $\Delta \Delta$, with mean-variance normalization over a moving 300 millisecond window). We add two models that have been shown to have representational spaces similar to human perceptual space \cite{cogsci2020}: multilingual bottleneck features (\textbf{Bot}:  \cite{bottleneck}) and a Dirichlet process Gaussian mixture model (\textbf{DP}: \cite{chen2015parallel}). \textbf{Bot} representations are from a bottleneck layer of a model trained to label phoneme states in seventeen phonetically diverse languages (107 hours training data in total; French and English are not included).\footnote{If the same sound belongs to different inventories, it is treated as distinct, for a total of 1032 possible phonemes.} We use the Shennong package\footnote{https://github.com/bootphon/shennong} to extract these representations. \textbf{DP} is a Gaussian mixture model over individual frames, from which we draw posteriorgrams. We use pretrained models from \cite{millet2019comparing}, trained on 34 hours of English speech and on 33 hours and 42 minutes of French speech, both from LibriVox.

For each ZeroSpeech 2017 model, we use the English and French features as they were submitted to the challenge, in the one second condition. All the models are presented in \cite{dunbar2017zero}, and we evaluate the systems [\textbf{H}] \cite{heck2017feature}, [\textbf{P1}], [\textbf{P2}], \cite{manenti2017unsupervised} [\textbf{A3}] \cite{ansari2017deep}, [\textbf{Y1}], [\textbf{Y2}] \cite{yuan2017extracting}, [\textbf{S1}], [\textbf{S2}] \cite{shibata2017composite}, [\textbf{C1}] and [\textbf{C2}] \cite{chen2017multilingual}. Broadly the systems follow four types of strategies. The first (\textbf{[H]}, \textbf{[P1]} and \textbf{[P2]}) consists in performing a bottom-up frame-level clustering. The second (\textbf{[C1]}, \textbf{[C2]} and \textbf{[A3]}) is to construct language-independent embeddings by training neural networks in an unsupervised way. The third (\textbf{[Y1]} and \textbf{[Y2]}) is to use spoken term discovery to improve the acoustic features. \textbf{[Y3]} is a supervised version of these systems, using transcribed pairs from the Switchboard data set. Finally, \textbf{[S1]} and \textbf{[S2]} use supervised pre-training on out-of-domain languages (only Japanese for \textbf{[S1]} and ten different languages, including French and English, for \textbf{[S2]}). Except for \textbf{[S1]}, \textbf{[S2]} and \textbf{[Y3]}, all the models are trained only on the 2017 challenge data. \textbf{[H]} and \textbf{[P1]} use only monolingual data, whereas the rest of the models, while trained separately for each language, make use of the complete set of  languages provided with the ZeroSpeech challenge during training. We followed the original choices of the authors concerning $\gamma$ (cosine distance for all, except for \textbf{[H]}, which uses symmetrised KL-divergence). We use the cosine distance for all reference models, except for \textbf{DP} and the topline, for which we use the symmetrised KL-divergence.

\begin{table*}[th]
\centering
  \caption{2017 ZeroSpeech challenge ABX \% accuracies (one-second across-speaker conditions), ABX \% accuracies on Perceptimatic, and ABX \% accuracy reweighted by human discriminability.  Higher scores are better.}
  \label{tab:errors}
  
  \begin{tabular}{|p{1.5cm}l|ccccccccccccccc|}
    \hline
 & &\scriptsize{topline} &   \footnotesize{S2}   & \footnotesize{Bot} & \footnotesize{H}  & \footnotesize{S1}   & \footnotesize{DP}   & \footnotesize{C2}   & \footnotesize{A3}    & \footnotesize{Y3}   & \footnotesize{C1}   & \footnotesize{Y2}   & \footnotesize{Y1}   & \footnotesize{P2}   & \footnotesize{P1}   & \scriptsize{MFCC} \\ \hline
\multirow{2}{0.75pt}{\footnotesize{Zerospeech 2017}} & \scriptsize{French} &\footnotesize{\textbf{89.4}}&\footnotesize{88.8} & \footnotesize{86.9}    & \footnotesize{86.4}   & \footnotesize{86.3} & \footnotesize{83.7} & \footnotesize{83} & \footnotesize{82.8} & \footnotesize{82.3} & \footnotesize{82.4} & \footnotesize{81.3} & \footnotesize{81.1} & \footnotesize{79.9} & \footnotesize{79.7} & \footnotesize{74.9} \\ & \scriptsize{English}&\footnotesize{91.4} &\footnotesize{\textbf{92.1}}  & \footnotesize{90.7}   & \footnotesize{89.9} & \footnotesize{89.9} & \footnotesize{88.8} & \footnotesize{87.3} & \footnotesize{86.8} & \footnotesize{86.4} & \footnotesize{86.2} & \footnotesize{86}   & \footnotesize{85.8} & \footnotesize{82.4} & \footnotesize{82.4} & \footnotesize{73.4} \\ \hline
\multirow{2}{0.75pt}{\footnotesize{Perceptimatic}} & \scriptsize{French}&\footnotesize{\textbf{92.8}}&\footnotesize{92.3} & \footnotesize{88.5}   & \footnotesize{88.6} & \footnotesize{89.6} & \footnotesize{86.5} & \footnotesize{85.8} & \footnotesize{84.4}& \footnotesize{84.5} & \footnotesize{84.7}& \footnotesize{83.3} & \footnotesize{83.4} & \footnotesize{81.7} & \footnotesize{81.9} & \footnotesize{78.3} \\ 
 & \scriptsize{English}&\footnotesize{91.8}&\footnotesize{\textbf{92}}  & \footnotesize{89.1}    & \footnotesize{90.6} & \footnotesize{88.5} & \footnotesize{88.7} & \footnotesize{86.9}& \footnotesize{82.8}& \footnotesize{84.9}& \footnotesize{84.5} & \footnotesize{83.2} & \footnotesize{82.7} & \footnotesize{82.5}& \footnotesize{81.8}& \footnotesize{78.2} \\ \hline
\multirow{2}{0.75pt}{\footnotesize{Perceptually Weighted}}& \scriptsize{French}&\footnotesize{93.4}&\footnotesize{\textbf{94}}& \footnotesize{90.1}   & \footnotesize{90.3}&   \footnotesize{91.3} & \footnotesize{88.5}&
\footnotesize{86.5} & \footnotesize{86.6}&\footnotesize{86.3}& \footnotesize{87.4}&  \footnotesize{85.1}&\footnotesize{85.2}  &\footnotesize{83.2}
&\footnotesize{83.2} & \footnotesize{79.9}\\
& \scriptsize{English}&\footnotesize{92.8}& \footnotesize{\textbf{94.1}} & \footnotesize{91.2}    & \footnotesize{93.0} & \footnotesize{91.1} & \footnotesize{91} 
& \footnotesize{87.6}&\footnotesize{85.7} & \footnotesize{87.1}& \footnotesize{89} &\footnotesize{85.7}& \footnotesize{85.1} &\footnotesize{84.8}

 & \footnotesize{84.3}& \footnotesize{80.6}\\ \hline
  \end{tabular}
\end{table*}

\section{Results}

Original ZeroSpeech 2017 ABX accuracies are given on the first row of Table \ref{tab:errors}. The original 2017 ZeroSpeech test set contains errors, partially corrected by our modifications (see Section \ref{sec:benchmark-stimuli}). ABX accuracies computed only on the Perceptimatic triplets are on the row \textbf{Perceptimatic}. We observe that the English and French scores become more similar (French scores increase and English scores decrease). Future submissions to ZeroSpeech 2017 should use our data set for a better evaluation of their model. We also re-weight the results of each model by humans' results (see row \textbf{Perceptually Weighted}) using \eqref{eq:rew}. 
We observe that re-weighting ABX accuracies by giving more importance to contrasts that are ``easy'' improves all the scores (see Table \ref{tab:errors}). This implies that globally, models already have more trouble with ``hard'' contrasts than with ``easy'' ones. 

We then evaluate models based on how well they predict human behaviour in the experiment. Each ``model'' we evaluate is in fact two trained models (one French, one English), which each make a gradient prediction of discriminability, $\delta$ for each triplet item. We use these two $\delta$ values as predictors for the response accuracy of the respective human native listeners (the French $\delta$ predicts French stimuli/listeners, and the English $\delta$ English stimuli/listeners) on each item. We then calculate the overall log-likelihood of a binary-response regression  with respect to the experimental data.  (See \ref{sec:method_comp} above.) This yields a measure of how well each model predicts human behaviour. To generate confidence intervals, we resample (N=13655: for each triplet item, we draw exactly three human responses without replacement) and re-fit the regressions.\footnote{Confidence intervals for the differences in log-likelihood between each pair of models can be found on the online repository.}


\begin{figure}[t]
  \centering
  \includegraphics[width=0.8\linewidth]{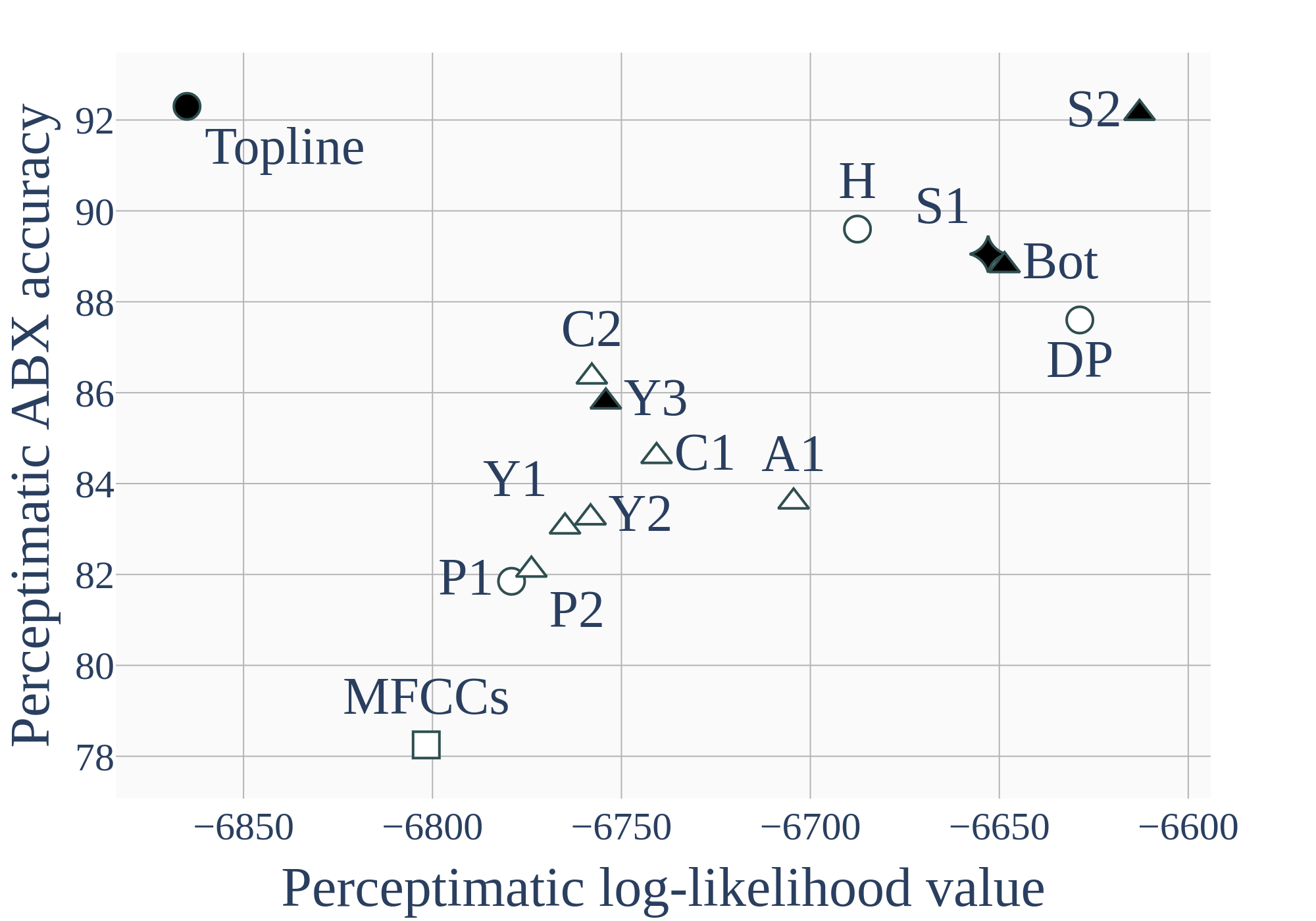}
  \caption{Perceptimatic ABX accuracy (mean of French and English: higher is better) versus log-likelihood. Higher log-likelihood values indicate representational spaces more similar to humans' perceptual space. White: unsupervised; black: supervised. Triangle: multilingual; circles: monolingual on  evaluated language; diamond: monolingual on another language, square: MFCCs. }
  \label{fig:results}
\end{figure}

Figure \ref{fig:results} plots the phone discrimination accuracy  (\textbf{Perceptimatic}) against the mean resampled log-likelihoods. We observe that having a representational space similar to humans' perceptual space is correlated with having a good ABX accuracy. However, we notice an obvious outlier: the (topline) supervised ASR model provided as a reference system for ZeroSpeech 2017. This model is unexpectedly bad at predicting human results. This result is consistent with results obtained for neural ASR models trained exclusively on the native language of human listeners \cite{cogsci2020}. Unsupervised models (such as \textbf{DP}) and multilingual models (such as \textbf{[S1]}, \textbf{[S2]}, and \textbf{Bot}) seem to be more predictive of human perception. It is therefore possible to discriminate phones well while making phoneme confusions very different from those of human beings. We discuss some possible explanations for this below. 

Conversely, the fact that \textbf{DP} obtains the second-highest log-likelihood (not significantly different from \textbf{[S2]}) is surprising, given its lower  discrimination accuracy. \textbf{[H]}, for example, built on \textbf{DP}  to generate a more speaker-invariant representation (it uses DPGMM clusters as the basis for supervised talker normalization techniques). While these improvements improve phone discriminability, they yield a less human-like representation. 

\begin{figure}[t]
  \centering
  \includegraphics[width=\linewidth]{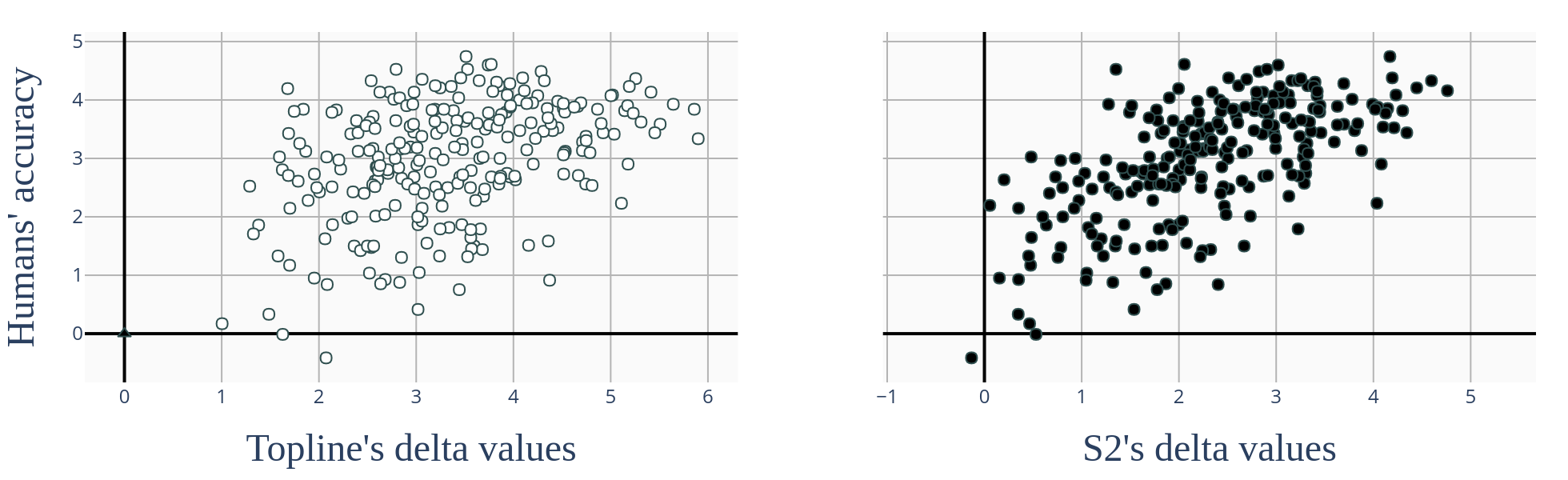}
  \caption{Average of French listeners' non-binarized results (higher: correct, and more certain of it) against average  $\delta$ from (\textbf{left}) supervised ASR (\textbf{right}) multilingual system \textbf{[S2]}. Each point is a phone pair. Measures are normalized by dividing by standard deviation over the entire data set.}
  \label{fig:results_plot}
\end{figure}

The best model, both in terms of perceptually-reweighted accuracy, and in terms of predicting human behaviour, is  \textbf{[S2]}. Similarly, \textbf{Bot} also performs well. These two models train on multilingual supervised phone recognition objectives. \textbf{[S2]} is then further adapted to either English or French, while  \textbf{Bot} is fixed (and thus has the same $\delta$ for the two stimuli languages). That these systems perform so well suggests that human behaviour in this task is driven mainly by universal, general speech perception mechanisms. With respect to this behaviour, monolingual supervised ASR may be over-training to the phone identification objective.  Figure \ref{fig:results_plot} plots French listeners' continuous discriminability scores against the French $\delta$ for supervised ASR and for \textbf{[S2]}, normalized to be  comparable. The ASR $\delta$ values are all shifted to the  right, indicating better discriminability, but at the expense of a poorer correlation with human accuracy.

\section{Overview}
We have presented Perceptimatic, a data set based on the 2017 ZeroSpeech challenge evaluation data. We used it to evaluate the models submitted to the 2017 ZeroSpeech challenge, and showed that, surprisingly, the supervised ASR reference model has a representational space very different from the perceptual space of human listeners. On the other hand, unsupervised models, and supervised models trained on other languages than the one being tested, seem to have more human-like representations. We leave open at least two possible explanations for this result. One is that human listening is simply not tuned to optimize phone classification in the native language, unlike supervised ASR. Another possibility rests on the observation that extracts of phones drawn from running speech, while standard for evaluating unsupervised speech representation learning, are very different from the types of clean speech stimuli used in typical human speech perception experiments. For humans, discriminating naturalistic examples of phones may tap into a more acoustic mode of listening; some other type of listening test might give results more comparable to a supervised ASR. 


\section{Acknowledgements}
This research was supported by the \'Ecole Doctorale Frontières du Vivant (FdV) -- Programme Bettencourt, and by grants ANR-17-CE28-0009 (GEOMPHON), ANR-11-IDFI-023 (IIFR), ANR-11-IDEX-0005 (USPC),  ANR-10-LABX-0083 (EFL), and ANR-17-EURE-0017.

\bibliographystyle{IEEEtran}

\bibliography{mybib}


\end{document}